\documentclass[10pt,english,final,a4paper,notitlepage]{article}
\pdfoutput=1

\usepackage[usenames,dvipsnames]{color}
\usepackage{hyperref} 
\hypersetup{
unicode=false,          
pdfauthor={JuanPi Carbajal, Cristiano Alessandro},%
pdftitle={Dynamic basis motion generation},%
colorlinks=true,       
linkcolor=OliveGreen,          
citecolor=Sepia,        
filecolor=magenta,      
urlcolor=NavyBlue,          
}

\usepackage[numbers,sort&compress]{natbib}

\usepackage{nicefrac}

\usepackage[utf8]{inputenc} 
\usepackage[T1]{fontenc}
\usepackage{textcomp}
\usepackage{lmodern} 

\usepackage{babel} 
\usepackage{graphicx}

\usepackage{amsmath}    
\usepackage{amsfonts}
\usepackage{amssymb}

\newcommand{\ud}{\mathrm{d}}

\DeclareMathOperator*{\argmin}{arg\,min}
\newcommand{\vps}[1]{\boldsymbol{#1}}
\newcommand{\vq}{\boldsymbol{q}}
\newcommand{\vdotq}{\boldsymbol{\dot{q}}}

\newcommand{\vu}{\boldsymbol{u}}
\newcommand{\vbq}[1]{\boldsymbol{#1{\theta}}}
\newcommand{\vBQ}{\boldsymbol{\Theta}}
\newcommand{\nBQ}{{N_\theta}}
\newcommand{\vba}{\boldsymbol{\phi}}
\newcommand{\vBA}{\boldsymbol{\Phi}}
\newcommand{\nBA}{{N_\phi}}

\newcommand{\err}{\operatorname{err}}

\newcommand{\mail}[1]{\href{mailto:#1}{#1}}
\begin{document}
\title{Synthesis and Adaptation of Effective Motor Synergies for the Solution of Reaching Tasks}
\author{Cristiano Alessandro \and Juan Pablo Carbajal \thanks{Artificial Intelligence Laboratory, University of Zürich, Switzerland. \mail{alessandro@ifi.uzh.ch}, \mail{carbajal@ifi.uzh.ch}}
 \and Andrea d'Avella\thanks{Laboratory of Neuromotor Physiology, Fondazione Santa Lucia, Italy. \mail{a.davella\@hsantalucia.it}}
}

\maketitle              

\begin{abstract}
Taking inspiration from the hypothesis of muscle synergies, we propose a method to generate open loop controllers for an agent solving point-to-point reaching tasks. The controller output is defined as a linear combination of a small set of predefined actuations, termed synergies. The method can be interpreted from a developmental perspective, since it allows the agent to autonomously synthesize and adapt an effective set of synergies to new behavioral needs. This scheme greatly reduces the dimensionality of the control problem, while keeping a good performance level. The framework is evaluated in a planar kinematic chain, and the quality of the solutions is quantified in several scenarios.
\end{abstract}

\section{Introduction}
Humans are able to perform a wide variety of tasks with great flexibility; learning new motions is relatively easy, and adapting to new situations (e.g. change in the environment or body growth) is usually dealt with no particular effort. The strategies adopted by the central nervous system (CNS) to master the complexity of the musculoskeletal apparatus and provide such performance are still not clear. However, it has been speculated that an underlying modular organization of the CNS may simplify the control and provide the observed adaptability. There is evidence that the muscle activity necessary to perform various tasks (e.g. running, walking, keeping balance, reaching and other combined movements) may emerge from the combination of predefined muscle patterns, the so-called \textit{muscle synergies} \citep{d'Avella2003}. This organization seems to explain muscle activity across a wide range of combined movements \citep{Ivanenko2005, Cappellini2006, D'Avella2008}.

The scheme of muscle synergies is inherently flexible and adaptable. Different actions are encoded by specific combinations of a small number of predefined synergies; this reduces the computational effort and the time required to learn new useful behaviors. The learning scheme can be regarded as developmental since information previously acquired (i.e. synergies) can be reused to generate new behaviors\citep{Dominici2011}. Finally, improved performance can be easily achieved by introducing additional synergies. Thus, the hypothetical scheme of muscle synergies would contribute to the autonomy and the flexibility observed in biological systems, and it could inspire new methods to endow artificial agents with such desirable features.

In this paper we propose a method to control a dynamical system (i.e. the agent) in point-to-point reaching tasks by linear combinations of a small set of predefined actuations (i.e. synergies). Our method initially solves the task in state variables by interpolation; then, it identifies the combination of synergies (i.e. actuation) that generate the closest kinematic trajectory to the computed interpolant. Additionally, we propose a strategy to synthesize a small set of synergies that is tailored to the task and the agent. The overall method can be interpreted in a developmental fashion; i.e. it allows the agent to autonomously synthesize and update its own synergies to increase the performance of new reaching tasks.

Other researchers in robotics and control engineering have recently proposed architectures inspired by the concept of muscle synergies. In \citep{Nori2005} the authors derive an analytical form of a set of primitives that can drive a feedback linearized system (known analytically) to any point of its configuration space. In \citep{Alessandro2012} the authors present a numerical method to identify synergies that optimally drive the system over a set of desired trajectories. This method does not require an analytical description of the system, and it has the advantage of assessing the quality of the synergies in task space. However, it is computationally expensive as it involves heavy optimizations. In \citep{Todorov2003} muscle synergies are identified by applying an unsupervised learning procedure to a collection of sensory-motor data obtained by actuating a robot with random signals. In \citep{Schaal05} the architecture of the dynamic movement primitives (DMP) is proposed as a novel tool to formalize control policies in terms of predefined differential equations. Linear combinations of Gaussian functions are used as inputs to modify the attractor landscapes of these equations, and to obtain the desired control policy.

In contrast to these works, our method to synthesize synergies does not rely on feedback linearization, nor on repeated integrations of the dynamical system. The method is grounded on the input-output relation of the dynamical system (as in \citep{Todorov2003}), and it provides a computationally fast method to obtain the synergy combinators to solve a given task. Furthermore, our method is inherently adaptable as it allows the on-line modification of the set of synergies to accommodate to new reaching tasks.

\section{Definitions and Methods\label{sec:methods}}
In this section we introduce the mathematical details of the method we propose. After some definitions, we present the core element of our method: a general procedure to compute actuations that solve point-to-point reaching tasks (see Sec. \protect\ref{sec:solTask}). Subsequently, in Section \protect\ref{sec:synthesis}, we propose a framework for the synthesis and the development of a set of synergies.

Let us consider a differential equation modeling a physical system\\* $\mathcal{D}\left(\vq(t)\right) = \vu(t)$, where $\vq(t)$ represents the time-evolution of its configuration variables (their derivatives with respect to time are $\vdotq(t)$), and $\vu(t)$ is the actuation applied. Inspired by the hypothesis of muscle synergies\footnote{With respect to the model of time-varying synergies, in this paper we neglect the synergy onset times.} \citep{d'Avella2003}, we formulate the actuation as a linear combination of predefined motor co-activation patterns:
\begin{equation}\label{exp:synergies}
  \vu(t) = \sum_{i=1}^{\nBA} \vba{}_i(t)b_i := \vBA(t)\vps{b},
\end{equation}

\noindent where the functions $\vba{}_i(t) \in \vBA$ are called \textit{motor synergies}. The notation $\vBA(t)$ describes a formal matrix where each column is a different synergy. If we consider a time discretization, $\vBA(t)$ becomes a $N\dim(\vq)$-by-$\nBA$ matrix, where $N$ is the number of time steps, $\dim(\vq)$ the dimension of the configuration space and $\nBA$ the number of synergies.

We define \textit{dynamic responses} (DR) of the set of synergies as the responses $\vbq{}_i(t)\in \vBQ$ of the system to each synergy (i.e. forward dynamics):
\begin{equation}
   \mathcal{D}(\vbq{}_i(t))=\vba{}_i(t) \hspace{5mm} i=1...\nBA.\label{exp:syngDR}
\end{equation}
\noindent with initial conditions chosen arbitrarily.

  \subsection{Solution to point-to-point reaching tasks\label{sec:solTask}}
  A general point-to-point reaching task consists in reaching a final state $\left(\vq_T,\vdotq_T\right)$ from an initial state $\left(\vq_0,\vdotq_0\right)$ in a given amount of time $T$:
  \begin{align}\label{exp:task}
  \begin{gathered}
    \vq(0) = \vq_0 , \quad \vdotq(0) = \vdotq_0, \\
    \vq(T) = \vq_T , \quad \vdotq(T) = \vdotq_T.
  \end{gathered}
  \end{align}

  \noindent Controlling a system to perform such tasks amounts to finding the actuation $\vu(t)$ that fulfills the point constraints\footnote{In this paper we assume that the initial conditions of the systems are equal to $\left(\vq_0,\vdotq_0\right)$} \eqref{exp:task}. Specifically, assuming that the synergies are known, the goal is to identify the appropriate synergy combinators $\vps{b}$. In this paper we consider only the subclass of reaching tasks that impose motionless initial and final postures, i.e. $\vdotq_T = \vdotq_0 = 0$.

  The procedure consists of, first, solving the problem in kinematic space (i.e. finding the appropriate $\vq(t)$), and then computing the corresponding actuations. From the kinematic point of view, the task can be seen as an interpolation problem; i.e. $\vq(t)$ is a function that interpolates the data in \eqref{exp:task}. Therefore, a set of functions is used to build the interpolant trajectory that satisfy the constraints imposed by the task; these functions are herein the dynamic responses of the synergies:
  \begin{equation}
    \vq(t) = \sum_{i=1}^{\nBQ} \vbq{}_i(t)a_i := \vBQ(t)\vps{a},\label{exp:solutionDR}
  \end{equation}
  \noindent where the vector of combinators $\vps{a}$ is chosen such that the task is solved. As mentioned earlier, if time is discretized, $\vBQ(t)$ becomes a $N\dim(\vq)$-by-$\nBQ$ matrix, where $\nBQ$ is the number of dynamic responses. The quality of the DR as interpolants is evaluated in sections \protect\ref{sec:results}.

  Once a kinematic solution has been found (as linear combination of DRs), the corresponding actuation can be obtained by applying the differential operator; i.e. $\mathcal{D}\left(\vBQ(t)\vps{a}\right) = \tilde{\vu}(t)$. Finally, the vector $\vps{b}$ can be computed by projecting $\tilde{\vu}(t)$ onto the synergy set $\vBA$. If $\tilde{\vu}(t)$ does not belong to the linear span of $\vBA$, the solution can only be approximated in terms of a defined norm (e.g. Euclidean):
  \begin{equation}
    \vps{b} = \argmin_{\vps{b}} ||\tilde{\vu}(t) - \vBA(t)\vps{b}||.\label{eq:minimization}
  \end{equation}
  \noindent When the time is discretized, all functions of time becomes vectors and this equation can be solved explicitly using the psuedoinverse of the matrix $\vBA$,
  \begin{equation}
    \vBA^{+}\tilde{\vu} = \vBA^{+}\mathcal{D}\left(\vBQ\vps{a}\right) = \vps{b}.
  \end{equation}
  \noindent This equation highlights the operator $\vBA^{+}\circ\mathcal{D}\circ\vBQ$ ($\circ$ denotes operator composition) as the mapping between the kinematic combinators $\vps{a}$ (kinematic solution) and the synergy combinators $\vps{b}$ (dynamic solution). Generically, this operator represents a nonlinear mapping $\mathcal{M}:\mathbb{R}^{N_\theta}\rightarrow \mathbb{R}^{N_\phi}$, and it will be discussed in Section \protect\ref{sec:discu}.\\*
  To assess the quality of the solution we define the following measures:\\*
  \emph{Interpolation error}: Measures the quality of the interpolant $\vBQ(t)\vps{a}$ with respect to the task. Strictly speaking, only the case of negligible errors corresponds to interpolation. A non-zero error indicates that the trajectory $\vBQ(t)\vps{a}$ only approximates the task
  \begin{equation}
  \err_I = \sqrt{||\vq_T - \vBQ(T)\vps{a}||^2 + ||\dot{\vBQ}(T)\vps{a}||^2},
  \end{equation}
  \noindent where $||\cdot||$ denotes the Euclidean norm, and the difference between angles are mapped to the interval $(-\pi,\pi]$.\\*
  \emph{Projection error}: Measures the distance between the actuation that solves the task $\tilde{\vps{u}}(t)$, and the linear span of the synergy set $\vBA$
  \begin{equation}
  \err_P = \sqrt{\int_0^T ||\tilde{\vu}(t) - \vBA(t)\vps{b}||^2 \ud t}.
  \end{equation}
  \noindent \emph{Forward dynamics error}: Measures the error of a trajectory $\tilde{\vps{q}}(t,\vps{\lambda})$ generated by an actuation $\vBA(t)\vps{\lambda}$, in relation to the task.
  \begin{equation}
  \err_F = \sqrt{||\tilde{\vq}(T,\vps{\lambda}) - \vq_T||^2 + ||\vps{\dot{\tilde{q}}}(T,\vps{\lambda}) - \vdotq_T||^2}.
  \end{equation}
  \noindent Replacing $\tilde{\vq}(t,\vps{\lambda})$, $\vq_T$ and $\vdotq_T$ with their corresponding end-effector values provides the \underline{forward dynamics error of the end-effector}.

  \subsection{Synthesis and Development of Synergies\label{sec:synthesis}}
The synthesis of synergies is carried on in two phases: exploration and reduction.   The exploration phase consists in actuating the system with an extensive set of motor signals $\vBA_0$ in order to obtain the corresponding DRs $\vBQ_0$. The reduction phase consists in solving a small number of point-to-point reaching tasks in kinematic space (that we call \textit{proto-tasks}) by creating the interpolants using the elements of set $\vBQ_0$, as described in Eq. \eqref{exp:solutionDR}. These solutions are then taken as the elements of the reduced set $\vBQ$. Finally, the synergy set $\vBA$ is computed using relation \eqref{exp:syngDR}, i.e. inverse dynamics. As a result, there will be as many synergies as the number of the proto-tasks (i.e. $\nBA = \nBQ$). The intuition behind this reduction is that the synergies that solve the proto-tasks may capture essential features both of the task and of the dynamics of the system. Despite the non-linearities of $\mathcal{D}$, linear combination of these synergies might be useful to solve point-to-point reaching tasks that are similar (in terms of Eq. \eqref{exp:task}) to the proto-tasks (see Sec. \protect\ref{sec:results}).

The number of proto-tasks as well as their specific instances determine the quality of the synergy-based controller. To obtain good performance in a wide variety of point-to-point reaching tasks, the proto-tasks should cover relevant regions of the state space (see Sec. \protect\ref{sec:results}). Clearly, the higher the number of different proto-tasks, the more regions that can be reached with good performance. However, a large number of proto-tasks (and the corresponding synergies) increases the dimensionality of the controller. In order to tackle this trade-off, we propose a procedure that parsimoniously adds a new proto-task only when and where it is needed: if the performance in a new reaching task is not satisfactory, we add a new proto-task in one of the regions with highest projection error or we modify existing ones.

\section{Results\label{sec:results}}
\begin{figure}[htbp]
\centering
\includegraphics[width=0.8\textwidth]{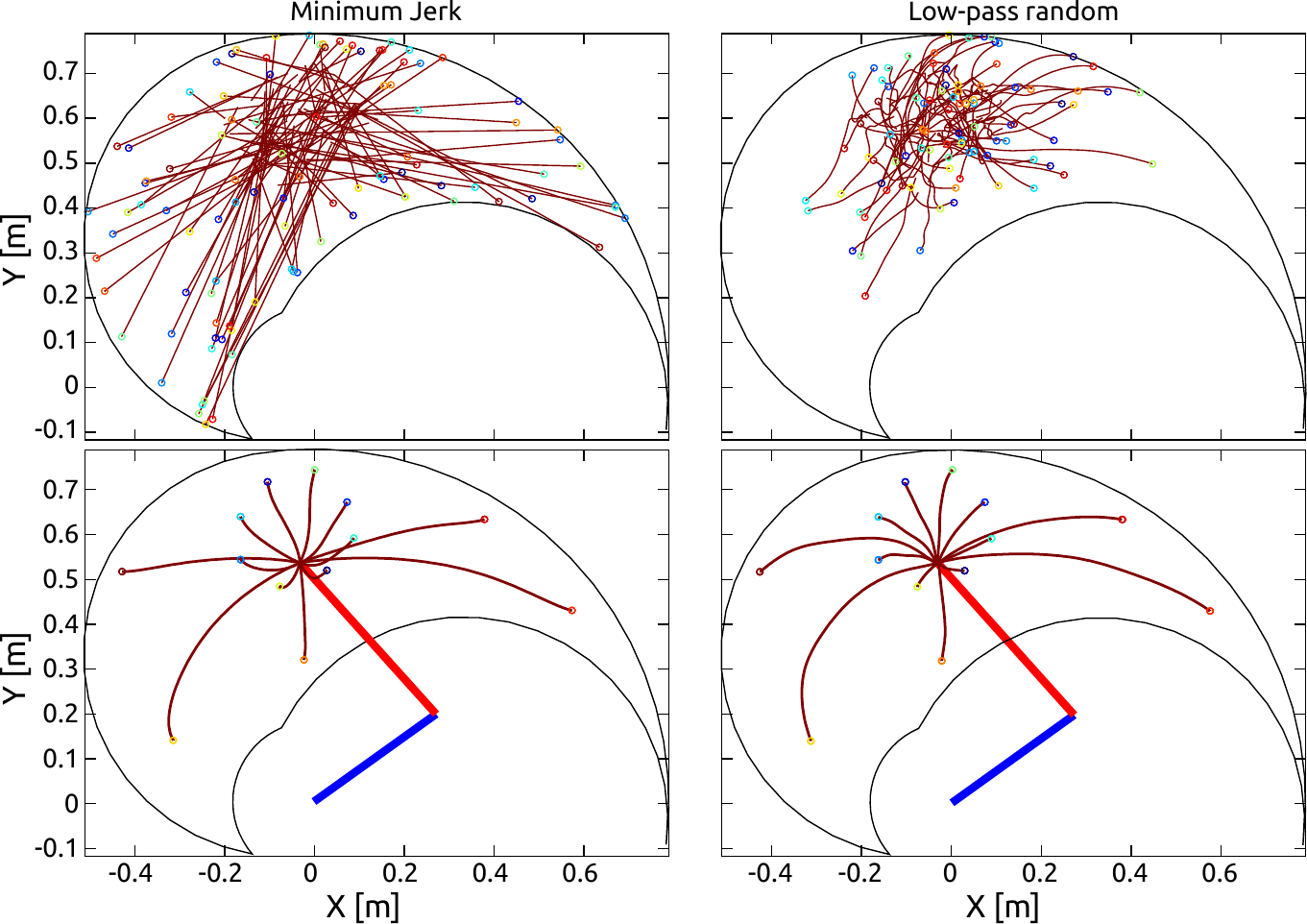}
\protect\caption{Comparison of explorations with two different classes of actuation: minimum jerk and low-passed random signal. Each panel shows the kinematic chain in it initial posture (straight segments). The limits of the end-effector are shown as the boundary in solid line.}
\label{fig:2DoF_Exploration}
\end{figure}

We apply the methodology described in Section \protect\ref{sec:methods} to a simulated planar kinematic chain (see \citep{Hollerbach1982} for model details) modeling a human arm\citep{Muceli2010}. In the exploration phase, we employ an extensive set of motor signals $\vBA_0$ to actuate the arm model and generate the corresponding dynamic responses $\vBQ_0$. The panels in the first row of Fig. \protect\ref{fig:2DoF_Exploration} show the end-effector trajectories resulting from the exploration phase. We test two different classes of motor signals: actuations that generate minimum jerk end-effector trajectories ($100$ signals), and low-passed uniformly random signals ($90$ signals). In order to evaluate the validity of the general method described in Sec. \protect\ref{sec:solTask}, we use the sets $\vBA_0$ and $\vBQ_0$ to solve $13$ different reaching tasks without performing the reduction phase. The second row of Fig. \protect\ref{fig:2DoF_Exploration} depicts the trajectories drawn by the end-effector when the computed mixture of synergies are applied as actuations (i.e. forward dynamics of the solution). It has to be noted how the nature of the solutions (as well as that of the responses), depends on the class of actuations used. The maximum errors are reported in Table \protect\ref{table:exploration_error}. The results are highly satisfactory for both the classes of actuations, and show the validity of the method proposed. Since the reduction phase has not been performed, the dimension of the combinator vectors $\vps{a}$ and $\vps{b}$ equals the number of actuations used in the exploration.
\begin{table}[hptb]
\centering
  \begin{tabular}{|c|cc|}
     \hline
	      & Min. Jerk 	& Random \\
     $\err_I$ & $10^{-15}$ 	& $10^{-15}$ \\\hline
     $\err_P$ & $10^{-5}$ 	& $10^{-3}$ \\\hline
     $\err_F$ & $10^{-4}$ 	& $10^{-3}$ \\\hline
  \end{tabular}
  \protect\caption{Order of the maximum errors obtained by using $\vBA_0$ and $\vBQ_0$ (no reduction phase).}
  \label{table:exploration_error}
\end{table}

\begin{figure}[htbp]
\centering
\includegraphics[width=0.85\textwidth]{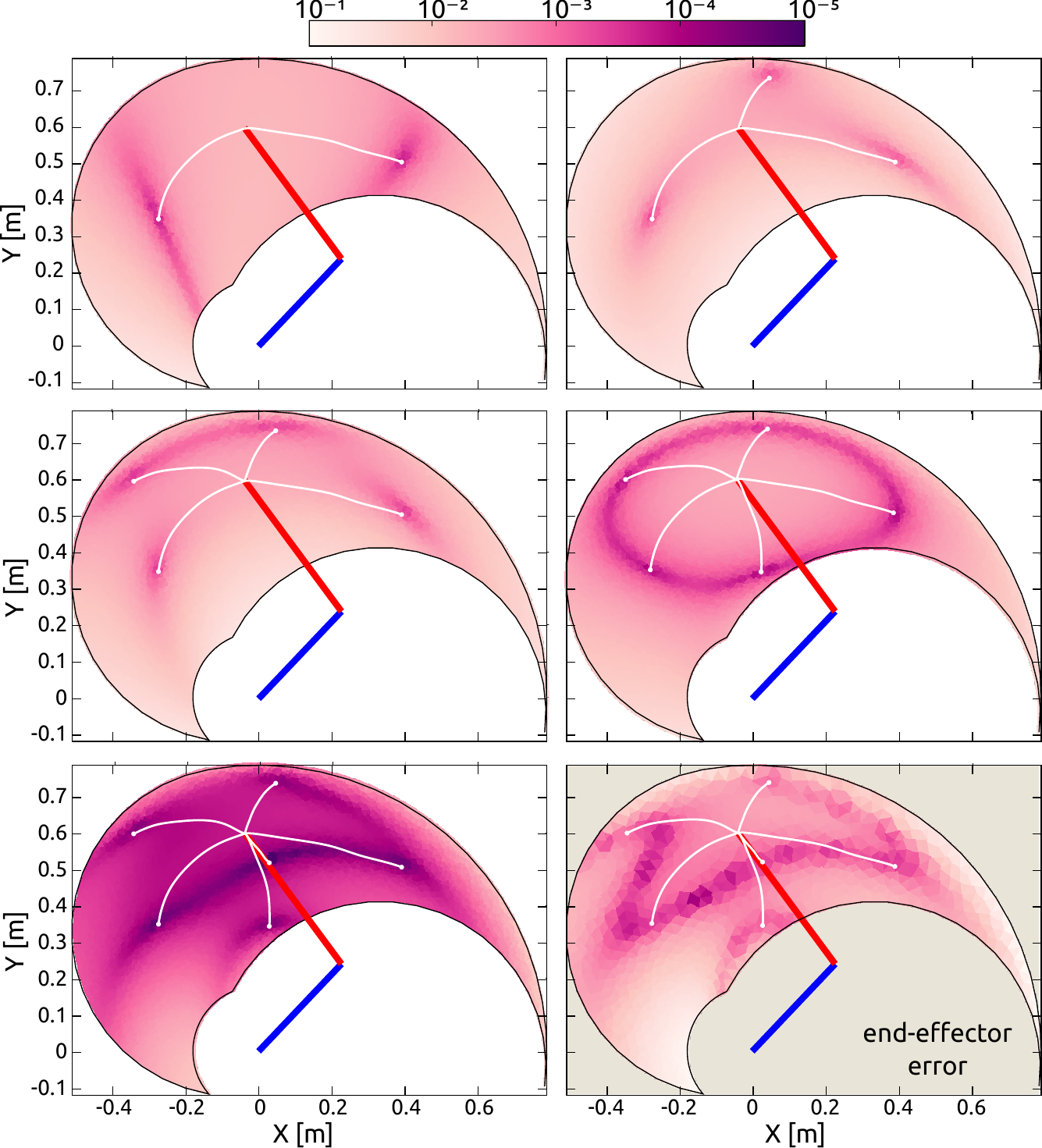}
\protect\caption{Selection of targets based on projection error. Each panel shows the kinematic chain in its initial posture (straight segments). The limits of the end-effector are the boundary of the colored regions. The color of each point indicates the projection error produced to reach a target in that position. The bottom right diagram shows the forward dynamics error of the end-effector using 6 proto-tasks (6 synergies).}
\label{fig:2DoF_Reduction}
\end{figure}

The objective of the reduction phase is to generate a small set of synergies and DRs that can solve desired reaching tasks effectively. As described in Section \protect\ref{sec:synthesis}, this is done by solving a handful of proto-tasks. The number (and the instances) of these proto-tasks determines the quality of the controller. Figure \protect\ref{fig:2DoF_Reduction} shows the projection error as a function of the number of proto-tasks. The reduction is applied to the low-passed random signal set. Initially, two targets are chosen randomly (top left panel); subsequent targets are then added on the regions characterized by higher projection error. As it can be seen, the introduction of new proto-tasks leads to better performance on wider regions of the end-effector space, and eventually the whole space can be reached with reasonable errors. In fact, the figure shows that this procedure decreases the average projection error to $10^{-3}$ (comparable to the performance of the whole set $\vBA_0$, see Tab. \protect\ref{table:exploration_error}) and reduces the dimension of the combinator vector to $6$, a fifteen-fold reduction. This result shows that a set of ``good'' synergies can drastically reduce the dimensionality of the controller, while maintaining similar performance. The bottom right panel of the figure shows the forward dynamics error of the end-effector obtained with the $6$ proto-tasks. Comparing this panel with the bottom left one, it can be seen that the forward dynamics error of the end-effector reproduces the distribution of the projection error, rendering the latter a good estimate for task performance.

\begin{figure}[htbp]
\centering
\includegraphics[width=\textwidth]{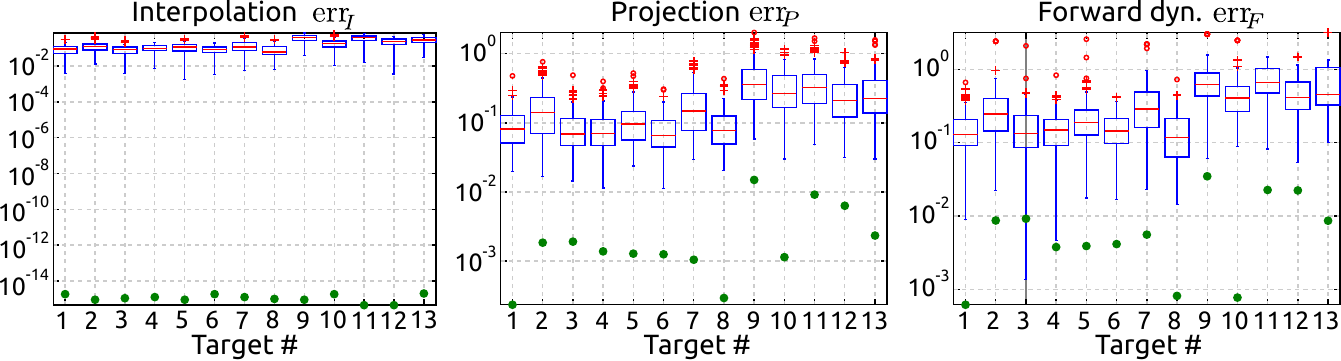}
\protect\caption{Evaluation of the reduction phase. Errors produced by subsets randomly selected from the exploration-actuations (boxplots) are compared with the errors obtained after the reduction phase (filled circles).}
\label{fig:2DoF_compareReduction}
\end{figure}

To further demonstrate that the reduction phase we propose is not trivial, we compare the errors resulting from the set of $6$ synthesized synergies, with the errors corresponding to $100$ random subsets of size $6$ drawn from the set of low-passed random motor signals. Figure \protect\ref{fig:2DoF_compareReduction} shows this comparison. The task consists in reaching the $13$ targets in Fig. \protect\ref{fig:2DoF_Exploration}. The boxplots correspond to the errors of the random subsets, and the filled circles to the errors of the synergies resulting from the reduction phase. Observe that, the order of the error of the reduced set is, in the worst case, equal to error of the best random subset. However, the mean error of the reduced set is about $2$ orders of magnitude lower. Therefore, the reduction by proto-tasks can produce a parsimonious set of synergies out of a extensive set of actuations. Evaluating the performance with different classes of proto-tasks (e.g. catching, hitting, via-points) is postponed to future works.


\section{Discussion\label{sec:discu}}
The results shown in the previous section justify the interpretation of the methodology as a developmental framework. Initially, the agent explores its sensory-motor system employing a variety of actuations. Later, it attempts to solve the first reaching tasks (proto-tasks), perhaps obtaining weak performance as the exploration phase may not have produced enough responses yet (see the box-plots in Fig. \protect\ref{fig:2DoF_compareReduction}). If the agent finds an acceptable solution to a proto-task, it is used to generate a new synergy (populating the set $\vBA$), otherwise it continues with the exploration. The failure to solve tasks of importance for its survival, could motivate the agent to include additional proto-tasks; Figure \protect\ref{fig:2DoF_Reduction} illustrates this mechanism. As it can be seen, the development of the synergy set incrementally improves the ability of the agent to perform point-to-point reaching. Alternatively, existing proto-tasks could be modified by means of a gradient descent or other learning algorithms. In a nutshell, the methodology we propose endows the agent with the ability to autonomously generate and update a set of synergies (and dynamic responses) that solve reaching tasks effectively.

Despite the difficulty of the mathematical problem (i.e nonlinear differential operator), our method seems to generate a small set of synergies that span the space of actuations required to solve reaching tasks. This is not a trivial result, since these synergies over-perform many other set of synergies randomly taken from the set $\vBA_0$ (see Fig. \protect\ref{fig:2DoF_compareReduction}). It appears as if the reduction phase builds features upon the exploration phase, that are necessary to solve new reaching tasks. To verify whether solving proto-tasks plays a fundamental role, our synergies could be compared with the principal components extracted from the exploration set. This verification goes beyond the scope of this paper.

An important aspect of our method is the relation between $\vBQ$ and $\vBA$ (see Eq. \eqref{exp:syngDR}). This mapping makes explicit use of the body parameters (embedded in the differential operator $\mathcal{D}$), hence the synergies obtained can always be realized as actuations. The same cannot be said, in general, for synergies identified from numerical analyses of biomechanical data. Though some studies have verified the feasibility of extracted synergies as actuations \citep{Neptune2009}, biomechanical constraints are not explicitly included in the extraction algorithms. Additionally, Eq. \eqref{exp:syngDR} provides an automatic way to cope with smooth variations of the morphology of the agent. That is, both the synergies and their dynamic responses evolve together with the body. In line with \cite{Nori2005, Alessandro2012}, these observations highlight the importance of the body in the hypothetical modularization of the CNS.

Once the task is solved in kinematic space, the corresponding actuation can be computed using the explicit inverse dynamical model of the system (i.e. the differential operator $\mathcal{D}$). It might appear that there is no particular advantage in projecting this solution onto the synergy set. However, the differential operator might be unknown. In this case, a synergy-based controller would allow to compute the appropriate actuation by evaluating the mapping $\mathcal{M}$ on the vector $\vps{a}$, hence obtaining the synergy combinators $\vps{b}$. Since $\mathcal{M}$ is a mapping between two finite low-dimensional vector spaces, estimating this map may turn to be easier than estimating the differential operator $\mathcal{D}$. Furthermore, we believe that the explicit use of $\mathcal{D}$ may harm the biological plausibility of our method. In order to estimate the map $\mathcal{M}$, the input-output data generated during the exploration phase (i.e. $\vBA_0$ and $\vBQ_0$) could be used as learning data-set. Further work is required to test these ideas. Additionally, preliminary theoretical considerations (not reported here) indicate that the synthesis of synergies without the explicit knowledge of $\mathcal{D}$ is also feasible.

Finally, the current formulation of the method does not includes joint limits explicitly. The interpolated trajectories are valid, i.e. they do not go beyond the limits, due to the lack of intricacy of the boundaries. In higher dimensions, especially when configuration space and end-effector are not mapped one-to-one, this may not be the case anymore. Nevertheless, joint limits can be included by reformulating the interpolation as a constrained minimization problem. Another solution might be the creation of proto-tasks with a tree-topology, relating our method to tree based path planning algorithms\citep{Shkolnik2011}.

\section{Conclusion and Future Work} \label{sec:conclusions}
The current work introduces a simple framework for the generation of open loop controllers based on synergies. The framework is applied to a planar kinematic chain to solve point-to-point reaching tasks. Synergies synthesized during the reduction phase over-perform hundreds of arbitrary choices of basic controllers taken from the exploration motor signals.
Furthermore, our results confirm that the introduction of new synergies increases the performance of reaching tasks. Overall, this shows that our method is able to generate effective synergies, greatly reducing the dimensionality of the problem, while keeping a good performance level. Additionally, the methodology offers a developmental interpretation of the emergence of task-related synergies that could be validated experimentally.

Due to the nonlinear nature of the operator $\mathcal{D}$, the theoretical grounding of the method poses a difficult challenge, and it is the focus of our current research. Another interesting line of investigation is the validation of our method against biological data, paving the way towards a predictive model for the hypothesis of muscle synergies. Similarly, the development of an automatic estimation process for the mapping $\mathcal{M}$ would further increase the biological plausibility of the model.

The inclusion of joint limits into the current formulation must be prioritized. Solving this problem will allow to test the method on higher dimensional redundant systems. Tree-based path planning algorithms may offer a computationally effective way to approach the issue.

\vspace{2mm}
\small{
\noindent{\bf Funds}: The research leading to these results has received funding from the European Community's Seventh Framework Programme FP7/2007-2013-Challenge 2 - Cognitive Systems, Interaction, Robotics- under grant agreement No 248311-AMARSi, and from the EU project RobotDoC under 235065 from the 7th Framework Programme (Marie Curie Action ITN).

\noindent{\bf Authors Contribution}: {\bf CA} and {\bf JPC} worked on the implementation of the algorithm and the generation of the results reported here. The method was born during {\bf JPC}'s visit to {\bf AD}'s laboratory. {\bf AD} provided material support for this development and uncountable conceptual inputs. All three authors have contributed to the creation of the manuscript. The authors list follows an alphabetical order.
}
%
%
\bibliographystyle{unsrt} 
\bibliography{references}

\end{document}